# Embedded Machine Learning for Microcontroller-Class Edge Devices: Data, Feature, Evaluation, and Deployment Pipelines

Mostafa Darvishi, *Senior Member*, *IEEE*
darvishi@ieee.org

***Abstract* - Embedded machine learning moves inference from cloud services to resource-constrained devices that must acquire data, preprocess signals, run a model, and act within tight limits on memory, energy, and latency. This paper presents a systems-oriented synthesis of an embedded machine-learning workflow for microcontroller-class platforms. The emphasis is placed on engineering decisions that are often hidden in generic machine-learning introductions: sampling and buffering, feature extraction as dimensionality reduction, validation under class imbalance, model/runtime co-design, and streaming deployment. Two representative signal families are used throughout the paper. The first is inertial motion recognition, where a two-second, three-axis accelerometer window is transformed from raw samples into root-mean-square and spectral features before classification. The second is keyword spotting, where audio is sampled, anti-aliased, transformed into mel-frequency cepstral coefficients, and processed by a compact one-dimensional convolutional network. The paper concludes with practical design rules for robust on-device inference, including data curation, quantization, thresholding, scheduling, and field monitoring.**

## INTRODUCTION

Machine-learning inference at the network edge is attractive when a sensor node must react immediately, preserve privacy, reduce communication energy, or operate without continuous connectivity. The engineering difficulty is that the target device is not a cloud server and often not even a single-board computer. A microcontroller may run bare metal or a small real-time operating system, have no virtual memory, expose only a few tens or hundreds of kilobytes of random-access memory (RAM), and execute from flash with a tight energy budget. Embedded machine learning is therefore best understood as a *co-design* problem among signal acquisition, feature representation, model architecture, runtime implementation, and application logic [1]-[4].

This paper aims to provide a compact technical synthesis of embedded machine learning concept targeting microcontrollers. Basic notions of artificial intelligence, such as what a classifier is or why a neural network has weights, are intentionally omitted. Instead, the discussion focuses on the steps that make an embedded model deployable: how raw signals become bounded input tensors, how validation metrics should be interpreted before deployment, and how the deployed program schedules acquisition, feature extraction, and inference without losing samples. The complete embedded machine-learning workflow considered in this paper is summarized in Fig. 1, where sensing, preprocessing, feature extraction, training, deployment, and field monitoring are treated as a closed-loop engineering process.

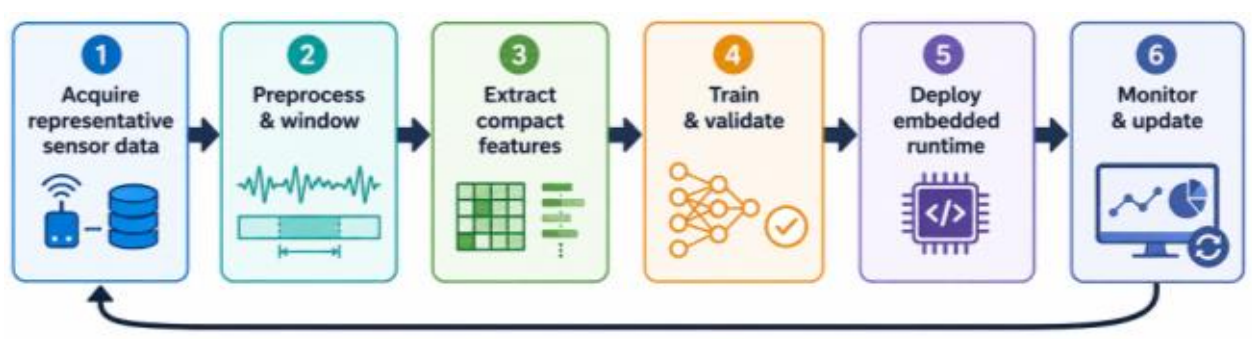

Fig. 1. Embedded machine-learning lifecycle. The deployed system is not the end of the workflow; field observations must feed back into data collection, feature selection, model revision, and runtime constraints.

The main contribution is a deployment-centered reference pipeline. It connects data curation, windowing, feature extraction, model training, evaluation, and runtime execution for two practical examples: inertial gesture recognition and small-footprint keyword spotting. The resulting workflow is not tied to a single tool. Edge Impulse, TensorFlow Lite Micro, CMSIS-NN, and similar frameworks can be viewed as implementation vehicles for the same architectural pattern [1], [3], [4].

## EMBEDDED TARGET MODEL

A central distinction in embedded machine learning is the difference between *single-board computers* and *microcontrollers*. A single-board computer can run a general-purpose operating system, provide command-line or graphical interfaces, and use a faster processor with more memory. A microcontroller is typically less powerful, but it is cheaper, lower power, and better suited to deterministic control loops, interrupt-driven sampling, and always-on sensing. The model designer must decide early whether the target can tolerate operating-system jitter, dynamic allocation, and higher power consumption, or whether the application requires a bare-metal or real-time design.

The embedded inference path also combines two modes of reasoning. Acquisition and control are *deterministic*: timers, direct memory access, interrupts, and real-time tasks must execute within known bounds. The trained model is *probabilistic*: it produces class probabilities or anomaly scores rather than deterministic truth. A reliable product must therefore surround the model with deterministic safeguards such as thresholds, debouncing, hysteresis, watchdogs, bounded execution time, and fallback states.

For a windowed sensor pipeline, the raw input size is $N_{raw} = f_s.T.C$, where $f_s$ is the sampling frequency, $T$ is the window length, and $C$ is the number of channels. This simple expression often determines whether a design can run on the chosen device. For example, 62.5 Hz sampling over a two-second, three-axis accelerometer window produces 375 raw values before feature extraction. If each value is stored as a 16-bit integer, the raw buffer alone requires 750 bytes, before double buffering, feature memory, model weights, activations, and runtime arena are considered.

## Data Curation and Windowing

Data collection is the first embedded design decision, not merely a machine-learning prerequisite. Samples should be acquired with the same sensor orientation, sample rate, quantization, environmental noise, and user behavior expected during deployment. A model trained from laboratory recordings can fail when microphone gain, sensor mounting, background motion, or device enclosure changes. For audio, sample matching includes sample rate, bit depth, and duration; for inertial data, it includes axis orientation, sampling jitter, gravity alignment, and mechanical coupling to the user or object.

The holdout method remains useful, but its embedded interpretation requires care. A common split such as 60 percent training, 20 percent validation, and 20 percent testing is meaningful only if the split prevents leakage across users, sessions, environments, or nearly duplicate windows. Randomly dividing neighboring windows from one long recording can overstate field performance because the validation set becomes too similar to the training set. For a device intended to generalize to new users or rooms, the validation and test sets should be separated by subject, environment, or capture session whenever possible.

Class balance and negative examples are especially important for edge devices. In keyword spotting, a deployed system must reject noise and unknown words, not only distinguish target words. In motion recognition, idle or out-of-distribution behavior can dominate real usage. Accuracy by itself can therefore be misleading: a naive classifier that predicts the majority class may appear successful while failing all minority events. Balanced data, representative unknown/noise classes, and anomaly detection are practical defenses against such failure modes.

## Feature Extraction as Embedded Compression

Feature extraction is a form of signal-informed compression. Instead of sending a high-dimensional raw window directly to a neural network, the embedded pipeline can compute features that preserve task-relevant structure while reducing memory and multiply-accumulate demand. This trade-off is not a retreat from deep learning; it is an engineering decision. For microcontroller-class devices, the cost of extracting features can be lower than the cost of running a larger end-to-end model, and the resulting feature vector may be easier to validate and debug.

In the inertial example, a two-second window contains 125 samples per axis and 375 raw values total. One compact representation computes root-mean-square (RMS) values and power spectral density (PSD) descriptors. With one root-mean-square value, three peak amplitudes, three peak frequencies, and four spectral-band powers per axis, the input becomes 33 features for all three axes. This reduces the model input dimension by more than an order of magnitude while preserving both amplitude and frequency information relevant to repetitive gestures. The corresponding windowed inertial inference pipeline is shown in Fig. 2. Raw three-axis accelerometer samples are accumulated into bounded buffers, transformed into compact RMS and PSD features, and then passed to a small classifier that outputs class probabilities for application-level decisions.

Audio keyword spotting illustrates a richer feature pipeline. The microphone signal must be band-limited before or around analog-to-digital conversion (ADC) to avoid aliasing. The sampling rate must exceed twice the highest relevant frequency component, and the practical voice band motivates the use of a kilohertz-scale sample rate rather than arbitrary sampling. A short-time Fourier transform or spectrogram exposes how energy evolves over time. Mel-frequency cepstral coefficients (MFCC) further compress this representation by applying mel-spaced filterbanks, logarithmic compression, and a discrete cosine transform. A one-second audio window can thus become a compact time-frequency matrix for a small convolutional classifier.

The streaming audio inference path used for keyword spotting is summarized in Fig. 3. In this pipeline, microphone samples are anti-aliased and digitized at a fixed rate, converted into a sliding-window MFCC representation, processed by compact one-dimensional convolution and pooling blocks, and finally mapped to class scores through a Softmax output layer.

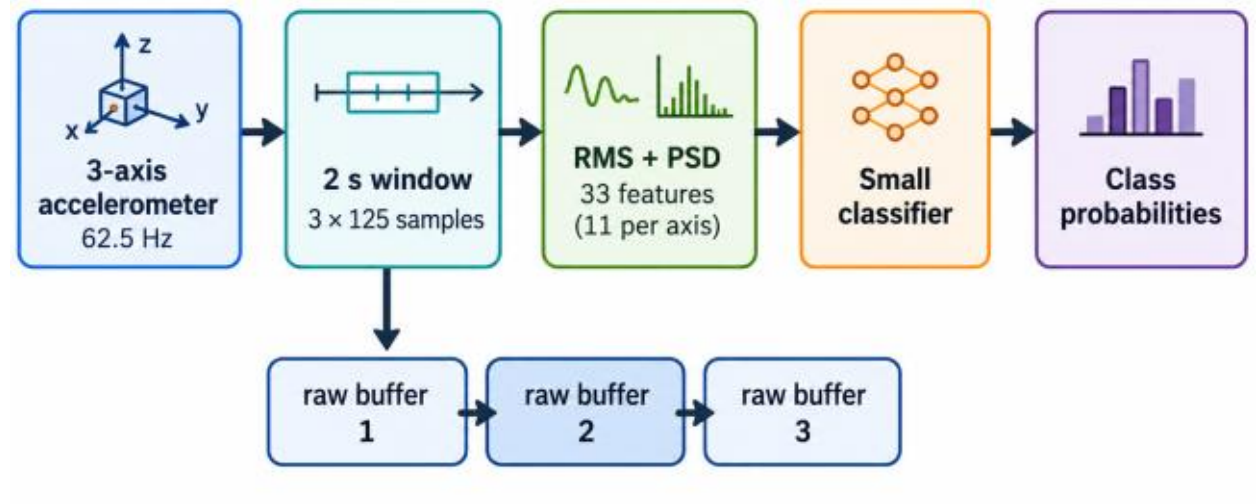

Fig. 2. Windowed inertial inference pipeline. Raw accelerometer samples are accumulated in bounded buffers, converted to spectral and amplitude features, classified, and converted into application decisions.

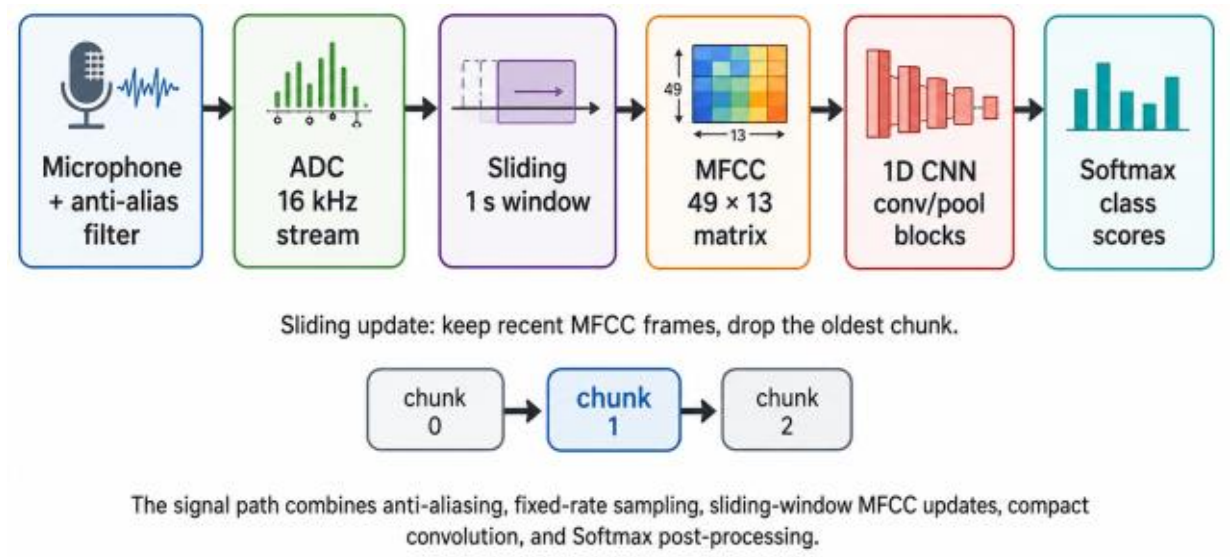

Fig. 3. Streaming keyword-spotting pipeline. the signal path combines anti-aliasing, fixed-rate sampling, sliding-window mfcc updates, compact convolution, and softmax post-processing.

## Model Architecture and Training

The appropriate model architecture depends on the representation. A handcrafted inertial feature vector can often be classified by a small fully connected network with a Softmax output. The Softmax values are not calibrated facts; they are model scores normalized to sum to one. The application must decide how scores are interpreted, for example by requiring $P_{(\text{left-right})} > 0.5$ or $P_{(\text{circle})} > 0.8$ before triggering an action. These thresholds are product parameters and should be evaluated with the same cost of false positives (FP) and false negatives (FN) expected in the field.

For audio, the MFCC matrix has local structure over time and cepstral index. One-dimensional convolution across the temporal axis is a natural compact architecture. A typical small-footprint pipeline reshapes the MFCCs into a two-dimensional array, applies one or more one-dimensional convolution and max-pooling blocks, flattens the representation, and ends with a Softmax over noise, unknown, and target keywords. Convolution exploits local patterns while sharing weights, which is useful when the number of parameters and operations must remain small [6].

Training behavior should be judged from both *loss* and *accuracy* curves. Underfitting occurs when both training and validation performance remain poor; useful remedies include more informative features, longer training, or a slightly more expressive model. Overfitting occurs when training performance improves while validation performance degrades; remedies include more data, early stopping, lower model complexity, regularization, dropout, or stronger data augmentation. Learning-rate diagnostics are equally important: a learning rate that is too low wastes training time, while one that is too high can produce unstable loss and poor convergence.

The representative configuration of embedded machine-learning models considered in this paper is summarized in Table I. The table emphasizes that model architecture is not selected only for predictive performance, but also for compatibility with the sensing modality, feature dimensionality, memory budget, and runtime execution model.

TABLE I. Representative embedded machine-learning model configuration

| Component | Motion model example | Audio model example | Embedded relevance |
|---|---|---|---|
| **Input** | 3-axis accelerometer | Microphone/audio stream | Sensor-driven inference |
| **Window** | 2 s, 3 × 125 samples | 1 s audio window | Bounded memory and latency |
| **Feature extraction** | RMS + PSD | MFCC | Reduces raw data dimensionality |
| **Classifier** | Small dense network | 1D CNN / compact classifier | Low memory footprint |
| **Output** | Class probabilities | Keyword probabilities | Application-level decision |
| **Deployment form** | TFLite / C++ runtime | TFLite / C++ runtime | MCU/SBC execution |

## Evaluation Metrics for Deployment

The confusion matrix should be treated as a deployment artifact. Each row corresponds to the actual class, each column to the predicted class, and the off-diagonal entries reveal product-specific failure modes. Total accuracy is useful only when class priors and costs are balanced. A model that rarely detects a target keyword but almost always predicts background can score well on an imbalanced test set while being unusable as a wake-word detector.

For a class $k$, the $F1$ score can be written as $F1.k = 2T.P.k / (2T.P.k + FP.k + FN.k)$. Unlike accuracy, this expression penalizes both missed detections and false alarms for the class of interest. Macro-averaged $F1$ is often more informative than total accuracy when the deployed data stream is dominated by idle, unknown, or background conditions. Per-class accuracy, precision, recall, false-positive rate (FPR), and false-negative rate (FNR) should be reviewed before threshold selection, especially if different classes trigger different actuation paths.

Evaluation should also include temporal behavior. A single correct window is not equivalent to a reliable event. Sliding-window classifiers can flicker between classes; keyword detectors can trigger on partial words; motion classifiers can output high confidence during transitions. Embedded products commonly need temporal smoothing, refractory intervals, and state machines around the model. These mechanisms should be evaluated along with the classifier rather than added after the fact.

The deployment-oriented evaluation criteria are summarized in Table II. These metrics show that embedded machine-learning evaluation must combine statistical performance with system-level feasibility, since a highly accurate model may still be unsuitable for deployment if it violates latency, memory, compute, or energy constraints.

TABLE II. EVALUATION METRICS RELEVANT TO EMBEDDED DEPLOYMENT

| Metric category | Metric | What it measures | Why it matters in embedded ML |
|---|---|---|---|
| **Predictive quality** | Accuracy | Overall correct classifications | General model correctness |
| **Predictive quality** | Precision, recall, F1 score | Class-level reliability | Important for imbalanced classes |
| **Error analysis** | Confusion matrix | Which classes are confused | Guides data/model improvement |
| **Runtime** | Latency | Time per inference | Real-time responsiveness |
| **Memory** | RAM / Flash usage | Runtime and storage footprint | Determines target compatibility |
| **Compute** | MACs / CPU cycles | Processing workload | Affects throughput and power |
| **Energy** | Energy per inference | Power cost of prediction | Critical for battery devices |

## EMBEDDED RUNTIME AND DEPLOYMENT

Deployment transforms a trained model into a bounded runtime. With TensorFlow Lite Micro, the model is converted into a FlatBuffer representation and linked with a C or C++ runtime. Many embedded runtimes use static memory allocation and an interpreter or generated code path rather than dynamic allocation. This matters because the model weights, input/output tensors, intermediate activations, operator scratch buffers, and runtime metadata must all fit within the target memory budget [3]. Optimized kernels such as CMSIS-NN can reduce inference latency and energy on Arm Cortex-M processors, but they do not remove the need for careful tensor sizing and operator selection [4].

The runtime schedule must avoid data loss. In the inertial example, the device samples every 16 ms to achieve 62.5 Hz. Sampling can be paced by a timer interrupt, an RTOS task, or direct memory access (DMA). The inference task should operate on a completed buffer while acquisition fills another buffer. If feature extraction and inference take longer than the stride between windows, the system either drops samples, increases latency, or must reduce model cost. This is why profiling on the actual hardware is more reliable than relying only on desktop model metrics.

The embedded deployment factors considered in this work are summarized in Table III. The table emphasizes that deployment is not only a matter of converting a trained model into an executable format, but also of ensuring deterministic data acquisition, bounded memory usage, efficient feature computation, stable runtime behavior, and reliable conversion of class probabilities into application-level decisions.

The application logic after inference is part of the model specification. A classifier that outputs $P_{(\text{left-right})} = 0.9143$, $P_{(\text{up-down})} = 0.0032$, $P_{(\text{circle})} = 0.0581$, and $P_{(\text{idle})} = 0.0244$ still needs a decision policy. The policy may select the maximum probability, apply class-specific thresholds, require repeated agreement across windows, or trigger an anomaly state if no class exceeds a minimum confidence. These rules should be documented, tested, and versioned with the model.

TABLE III. EMBEDDED RUNTIME AND DEPLOYMENT CONSIDERATIONS.

| Deployment aspect | Design consideration | Embedded implication |
|---|---|---|
| **Model format** | TFLite / FlatBuffer / generated C++ | Determines portability and runtime integration |
| **Memory allocation** | Static buffers preferred | Reduces fragmentation and improves determinism |
| **Sensor acquisition** | Timer interrupt, RTOS task, or DMA | Maintains fixed sampling rate |
| **Window management** | Rolling or double/triple buffers | Allows inference on completed windows |
| **Feature extraction** | RMS/PSD or MFCC computed on-device | Reduces model input size |
| **Inference runtime** | Interpreter or compiled model | Affects latency, memory, and portability |
| **Decision logic** | Probability thresholds / class scores | Converts model output to application action |
| **Field update** | Retraining or model replacement | Supports long-term robustness |

Quantization is usually necessary for microcontroller deployment. Integer-only inference reduces memory footprint and improves compatibility with integer hardware [8]. However, quantization changes numerical behavior, so calibration data and post-quantization validation are required. The final acceptance criterion should include flash size, peak random-access memory, mean and worst-case inference latency, energy per inference, and task-level accuracy under realistic noise and sampling conditions.

## CASE STUDY PATTERNS

### *I. Inertial gesture recognition*

The motion-recognition case study is representative of many embedded sensor tasks. Raw acceleration is cheap to acquire, but a raw two-second window can be unnecessarily large for a small classifier. RMS and PSD features reduce input dimensionality, expose periodic motion, and provide a stable input vector. A compact classifier can then distinguish *left-right*, *up-down*, *circle*, and *idle* classes. The key deployment risks are orientation mismatch, sensor saturation, inconsistent gesture speed, and transition windows between actions. These risks should be addressed by data augmentation, subject/session-aware validation, and state-machine filtering.

### *II. Small-footprint keyword spotting*

The keyword-spotting case study emphasizes the importance of signal processing before the neural network. Data curation must include target words, unknown words, and background noise. Augmentation can mix target utterances with background recordings to simulate deployment conditions. Sample matching is essential: changing sample rate, bit depth, or window length changes the input distribution. MFCCs provide a compact representation, and one-dimensional convolution extracts local temporal patterns with fewer parameters than a dense model applied to the flattened matrix [5], [6]. Streaming inference can update the MFCC matrix by dropping the oldest chunk and inserting the newest chunk, reducing repeated computation.

TABLE IV. DEPLOYMENT PATTERNS DERIVED FROM REPRESENTATIVE EMBEDDED MACHINE-LEARNING CASE STUDIES

| Pattern | Basis in the paper | Embedded ML lesson |
|---|---|---|
| **Windowed motion recognition** | Detailed inertial example | Fixed-size windows bound memory and latency |
| **Streaming keyword spotting** | Detailed audio/MFCC example | Sliding updates support continuous inference |
| **Sensor-paced acquisition** | Derived from both examples | Sampling must be synchronized with runtime execution |
| **Feature compression before inference** | Derived from RMS/PSD and MFCC examples | Compact features reduce memory and compute |
| **Probability-to-action logic** | Derived from classifier outputs | Softmax scores must be converted into deterministic decisions |
| **Field monitoring and update** | Derived from closed-loop workflow | Deployment requires feedback for drift and robustness |

The inertial motion-recognition and audio keyword-spotting examples illustrate two different sensing modalities, but they reveal a common embedded machine-learning structure. In both cases, deployment is governed not only by model accuracy, but also by fixed-rate acquisition, bounded windowing, compact feature extraction, lightweight inference, and deterministic conversion of class probabilities into application-level decisions. Table IV summarizes these deployment patterns and distinguishes the detailed case studies discussed in the paper from the broader engineering lessons derived from them.

## PRACTICAL DESIGN RULES

Embedded machine-learning design should begin with device and deployment constraints rather than with a neural-network architecture. The following rules summarize the main practical considerations for moving from an offline trained model to a reliable on-device inference system.

1. **Start with a resource and timing budget.** The design should first define the available flash memory, random-access memory (RAM), sensor sampling rate, window length, stride, maximum acceptable latency, energy per inference, and expected class priors. These values determine whether raw end-to-end learning is feasible or whether compact feature extraction followed by a small classifier is more appropriate.
2. **Treat the dataset as a product test artifact.** The training dataset should reflect the actual deployment configuration, including sensor mounting, enclosure effects, sensor drivers, preprocessing code, and sampling settings expected in the field. Negative examples should be curated as carefully as positive examples, since false activations can dominate real embedded behavior.
3. **Validate across realistic sources of variation.** The validation split should challenge the model with new sessions, users, devices, placements, or environments rather than only random samples from the same recording conditions. The final test set should be used once for final evaluation and should not be repeatedly reused for hyperparameter tuning.
4. **Design the runtime as a real-time embedded subsystem.** The deployed inference path should use bounded buffers, deterministic acquisition, and static memory allocation whenever possible. Inference latency should be profiled on the target hardware under realistic clock, compiler, and power settings rather than estimated only from desktop training results.
5. **Version the complete deployed pipeline.** A model update should not be versioned alone. The deployed artifact should include the model, preprocessing constants, feature-extraction parameters, quantization settings, probability thresholds, and post-processing logic. This is essential because small changes in preprocessing or thresholds can significantly alter field behavior.
6. **Evaluate predictive and deployment metrics together.** Accuracy, precision, recall, $F1$ score, and confusion-matrix behavior should be interpreted together with latency, memory footprint, processor utilization, and energy per inference. A model with high offline accuracy may still be unsuitable if it exceeds the available memory or violates the real-time response budget.
7. **Plan field monitoring before deployment.** Sensor aging, firmware changes, user behavior, and environmental noise can shift the deployed data distribution over time. Privacy-preserving logging of aggregate confidence, rejection rates, event counts, and failure modes can reveal when retraining or recalibration is required without necessarily storing raw personal data.
8. **Match on-device inference with privacy and reliability safeguards.**

Always-on audio, wearable, and health-related embedded systems benefit from local inference because raw data can remain on the device. However, this advantage should be paired with explicit safeguards for data minimization, robust decision thresholds, failure handling, and update traceability.

## Open Research Directions

The strongest future embedded machine-learning systems will combine algorithmic compression with hardware-aware scheduling. Neural architecture search, quantization-aware training, pruning, operator fusion, and tiny inference engines have shown that learning-based inference can be moved onto devices with memory and compute budgets far below those of mobile phones [9]. However, the next generation of embedded machine-learning systems will require more than smaller neural networks. The complete sensing-to-decision path must be co-designed so that acquisition, preprocessing, feature extraction, model execution, post-processing, and system-level response are optimized together under latency, memory, and energy constraints.

Several open problems remain particularly important for deployment. Continual learning is still difficult on constrained devices because adaptation must occur without excessive memory use, catastrophic forgetting, or uncontrolled changes in field behavior. Uncertainty calibration and out-of-distribution detection are also essential, since embedded systems frequently operate under sensor noise, mounting variation, environmental drift, and user-dependent behavior. In these cases, a model should not only predict a class, but also indicate when the input is unreliable or outside the conditions represented in the training data. Verification is another open challenge. Unlike conventional embedded software, machine-learning models are probabilistic, data-dependent components whose behavior cannot be fully described by deterministic control-flow analysis. Methods are therefore needed to verify threshold logic, rejection behavior, confidence margins, and safety-relevant decisions at the system level.

Another important direction is application-specific acceleration. Many embedded deployments do not spend all of their time in dense matrix multiplication. Filtering, window management, spectral analysis, MFCC computation, normalization, buffering, and post-processing may contribute significantly to total latency and energy. Future accelerators should therefore support complete embedded inference pipelines rather than only accelerating multiply-accumulate operations. In some applications, fusing feature extraction with inference may reduce memory traffic more effectively than increasing peak arithmetic throughput.

Scientific reporting should also improve. Many embedded model demonstrations report only accuracy, leaving out memory use, latency, energy, sampling assumptions, preprocessing cost, and failure modes. A deployable paper should report at least the target device, clock frequency, numerical precision, model size, peak memory, operator set, feature-extraction cost, end-to-end latency, class-wise metrics, and test conditions. For streaming systems, the sampling rate, window length, stride, buffering method, and decision policy should also be stated explicitly. Without these details, a model may appear accurate in an offline notebook while being impossible to reproduce on the embedded device it claims to target.

Finally, privacy-preserving and maintainable deployment remains a central research direction. On-device inference can reduce the need to transmit raw audio, health, motion, or environmental data, but privacy benefits depend on careful system design. Future systems should support data minimization, aggregate field monitoring, secure model updates, traceable dataset versions, and transparent reporting of known limitations. Embedded machine learning will therefore mature not only through better models, but through more rigorous integration of machine learning with embedded systems engineering.

## Conclusion

Embedded machine learning is not only the act of placing a trained model on a small device. It is a complete engineering pipeline that begins with representative sensing and ends with bounded, monitored, and maintainable inference. The most important design choices are often outside the neural network: sampling rate, window length, buffering, feature extraction, validation split, thresholding, and runtime scheduling. By focusing on motion recognition and keyword spotting, this paper has shown how raw sensor streams can be transformed into compact features, evaluated with class-sensitive and deployment-aware metrics, and deployed through deterministic embedded software. The same design pattern generalizes to many edge applications in health monitoring, industrial sensing, human-machine interfaces, and low-power intelligent devices. Future progress in this field will therefore depend not only on smaller models, but also on more rigorous co-design between data collection, feature representation, runtime execution, hardware constraints, and long-term field reliability. Thus, embedded machine learning should be treated as a system-level design discipline, not merely as model deployment.